\documentclass{article}

\usepackage[preprint]{colm2026_conference}
\usepackage{lineno}
\usepackage{latexsym}
\usepackage{amsmath}
\usepackage{amssymb}
\usepackage{graphicx}
\usepackage[export]{adjustbox}
\usepackage{booktabs}
\usepackage{multirow}
\usepackage[table]{xcolor}
\definecolor{TableSeparator}{gray}{0.92}
\definecolor{clmblue}{HTML}{E76F51}   % coral (ours)
\definecolor{mlmred}{HTML}{264653}    % dark teal (baseline)
\definecolor{freezegray}{HTML}{A8A8A8} % neutral gray
\definecolor{accentgreen}{HTML}{228833}
\definecolor{clmpurple}{HTML}{332288}
\definecolor{accentteal}{HTML}{44AA99}
\definecolor{phasebg}{HTML}{F0F4F8}
\definecolor{decaybg}{HTML}{FFF8F0}
\usepackage{siunitx}
\usepackage{float}
\usepackage{threeparttable}
\usepackage{tikz}
\usetikzlibrary{positioning,shapes,arrows,calc,decorations.pathreplacing}

\usepackage[colorlinks=true,linkcolor=blue,citecolor=blue,urlcolor=blue]{hyperref}

\title{A Causal Language Modeling Detour Improves Encoder Continued Pretraining}

\author{
Rian Touchent \\
Sorbonne Universit\'e / INRIA Paris \\
ALMAnaCH Team \\
\texttt{rian.touchent@inria.fr}
\And
\'Eric de la Clergerie \\
INRIA Paris \\
ALMAnaCH Team \\
\texttt{eric.de\_la\_clergerie@inria.fr}
}

\begin{document}
\ifcolmsubmission \linenumbers \fi

\maketitle

% ============================================================================
% ABSTRACT
% ============================================================================
\begin{abstract}
When adapting an encoder to a new domain, the standard approach is to continue training with Masked Language Modeling (MLM). We show that temporarily switching to Causal Language Modeling (CLM) followed by a short MLM decay improves downstream performance. On biomedical texts with ModernBERT, this CLM detour outperforms MLM baselines trained on identical data and compute across 8 French and 11 English biomedical tasks, by +1.2--2.8pp and +0.3--0.8pp respectively, depending on model size. We investigate the reasons for these gains. We find that CLM's dense supervision impacts low transformer layers (0--7) far more than MLM does. Freezing low layers during CLM eliminates the downstream benefit; freezing mid layers preserves it. The representational changes persist through the MLM decay phase, even when it matches the CLM phase in length, and they scale with model capacity. We release ModernCamemBERT-bio and ModernBERT-bio as state-of-the-art biomedical encoders in Base and Large sizes.
\end{abstract}

\newcommand{\hflogo}{\raisebox{-0.18em}{\includegraphics[height=1em]{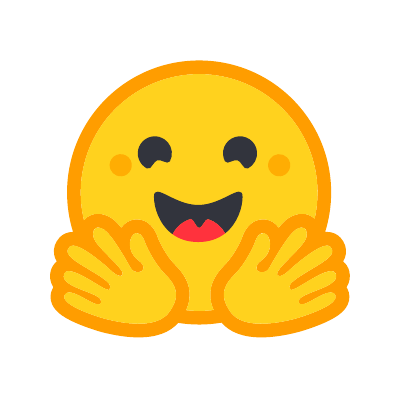}}}
\newcommand{\hfsize}[2]{\mbox{\href{https://huggingface.co/almanach/#1}{\texttt{#2}}}}
\newcommand{\hfmodel}[1]{\mbox{\href{https://huggingface.co/almanach/#1}{\texttt{almanach/#1}}}}
% --- Variant A (compact, cliquable base/large) ---
\vspace{0.8em}
\begin{flushleft}
\hspace{1em}\hflogo\,\textbf{ModernBERT-bio} (English):\quad
  \hfsize{ModernBERT-bio-base}{base}\quad
  \hfsize{ModernBERT-bio-large}{large} \\[0.35em]
\hspace{1em}\hflogo\,\textbf{ModernCamemBERT-bio} (French):\quad
  \hfsize{ModernCamemBERT-bio-base}{base}\quad
  \hfsize{ModernCamemBERT-bio-large}{large}
\end{flushleft}
\vspace{0.4em}
% --- Variant B (noms complets, aligné gauche, 1 modèle par ligne) ---
% \vspace{0.8em}
% \begin{flushleft}
% \hspace{1em}\begin{tabular}{@{}l@{\hspace{1.2em}}l@{}}
% \multirow{2}{*}{\hflogo\,\textbf{English}}
%   & \hfmodel{ModernBERT-bio-base} \\
%   & \hfmodel{ModernBERT-bio-large} \\[0.35em]
% \multirow{2}{*}{\hflogo\,\textbf{French}}
%   & \hfmodel{ModernCamemBERT-bio-base} \\
%   & \hfmodel{ModernCamemBERT-bio-large}
% \end{tabular}
% \end{flushleft}
% \vspace{0.4em}

% ============================================================================
% MAIN SECTIONS
% ============================================================================
\section{Introduction}
\label{sec:introduction}

\begin{figure*}[t]
\centering

\begin{minipage}[c]{0.45\textwidth}
\centering
\resizebox{\textwidth}{!}{%
\begin{tikzpicture}[
    font=\small,
    box/.style={draw, rounded corners=3pt, minimum height=0.7cm, font=\small\bfseries, line width=1.2pt},
    >=stealth,
]

\node[font=\normalsize\bfseries, anchor=west] at (0, 3.4) {(a) Pipeline};

% Phase labels (braces spanning both rows)
\draw[decorate, decoration={brace, amplitude=4pt, raise=2pt}] (2.95, 2.45) -- (4.45, 2.45)
    node[midway, above=6pt, font=\scriptsize] {CPT};
\draw[decorate, decoration={brace, amplitude=4pt, raise=2pt}] (5.45, 2.45) -- (6.95, 2.45)
    node[midway, above=6pt, font=\scriptsize] {Decay};

% MLM baseline (top row): Encoder -> MLM (CPT) -> MLM (Decay)
\node[box, draw=freezegray, fill=freezegray!15, minimum width=1.5cm] (e2) at (1.0, 1.9) {Encoder};
\node[box, draw=mlmred, fill=mlmred!30, minimum width=1.5cm] (m2a) at (3.7, 1.9) {MLM};
\node[box, draw=mlmred, fill=mlmred!30, minimum width=1.5cm] (m2b) at (6.2, 1.9) {MLM};
\draw[->, line width=1.6pt, mlmred] (e2) -- node[above, font=\footnotesize\bfseries] {10B} (m2a);
\draw[->, line width=1.6pt, mlmred] (m2a) -- node[above, font=\footnotesize\bfseries] {1B} (m2b);
\node[font=\footnotesize\bfseries, anchor=east] at (-0.1, 1.9) {Baseline};

% CLM detour (bottom row): Encoder -> CLM -> MLM
\node[box, draw=freezegray, fill=freezegray!15, minimum width=1.5cm] (e1) at (1.0, 0.5) {Encoder};
\node[box, draw=clmblue, fill=clmblue!35, minimum width=1.5cm] (c1) at (3.7, 0.5) {CLM};
\node[box, draw=mlmred, fill=mlmred!30, minimum width=1.5cm] (d1) at (6.2, 0.5) {MLM};
\draw[->, line width=1.6pt, clmblue] (e1) -- node[above, font=\footnotesize\bfseries] {10B} (c1);
\draw[->, line width=1.6pt, mlmred] (c1) -- node[above, font=\footnotesize\bfseries] {1B} (d1);
\node[font=\footnotesize\bfseries, anchor=east] at (-0.1, 0.5) {Ours};

%% ERIC: added a brace for detour
\draw[decorate, clmblue, decoration={brace, amplitude=4pt, raise=2pt, mirror}] (2.95,0) -- (4.45,0)
    node[midway, below=6pt, font=\scriptsize] {detour};

\end{tikzpicture}%
}% end resizebox
\end{minipage}%
\hfill
\begin{minipage}[c]{0.50\textwidth}
\centering
\vspace{-0.3cm}
\includegraphics[width=\textwidth]{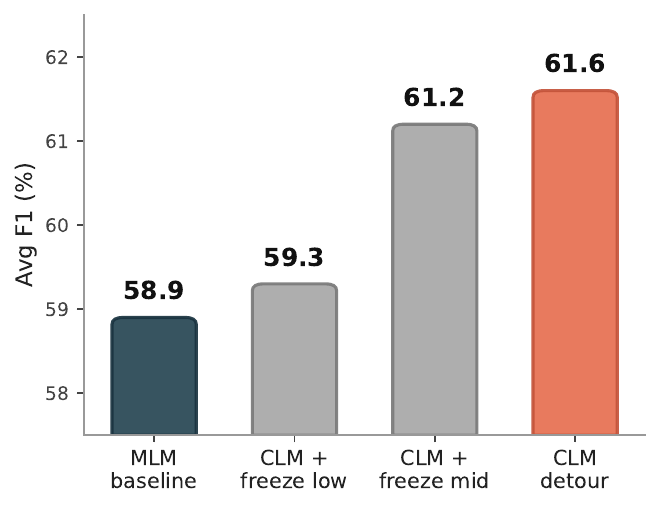}
\end{minipage}

\caption{\textbf{(a)} The CLM detour: a pretrained encoder trains with CLM, then returns to MLM (10\% decay). The MLM baseline trains with MLM throughout for matched compute. \textbf{(b)} Freeze interventions (French, 8 tasks, 9 seeds). Freezing low layers (0--7) during CLM detour drops performance to MLM baseline level; freezing mid layers (8--14) preserves the CLM benefit.}
\label{fig:overview}
\end{figure*}
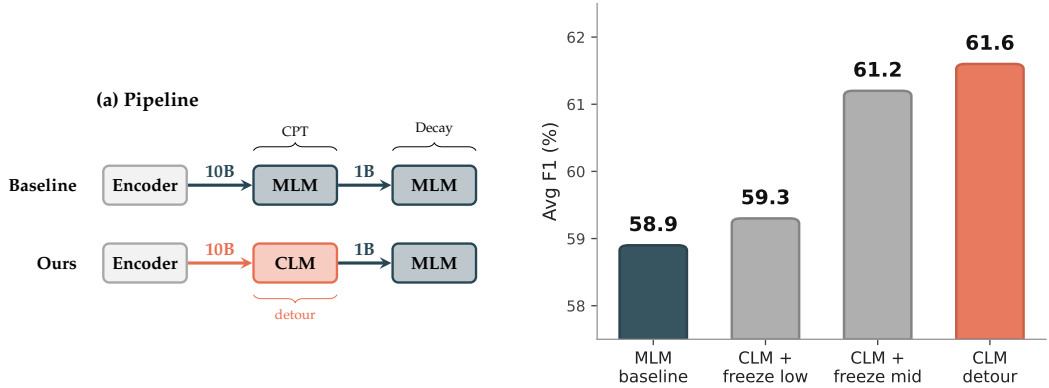

Domain-adaptive continued pretraining extends general-purpose language models to specialized domains \citep{gururangan2020dont,ke2022continual}. For encoders, this typically means extending masked language modeling (MLM) on domain text. We find that temporarily switching to causal language modeling (CLM) before returning to MLM, a \emph{CLM detour} (see Figure~\ref{fig:overview}), outperforms standard MLM continued pretraining on biomedical text, with the largest gains when the domain gap between pretraining and target data is large. The recipe changes only the attention mask and training objective, not the model architecture: use a causal mask for the CLM phase, then restore bidirectional attention and MLM for a short decay phase. With ModernBERT \citep{warner2024modernbert}, this produces state-of-the-art biomedical encoders in both English and French with 8,192-token context.

Yet the final model uses bidirectional attention and never performs CLM at inference. Why does a temporary objective switch leave a lasting benefit?

Comparing layer-by-layer representations between CLM-detour and MLM-only models with CKA \citep{kornblith2019similarity}, we observe that the CLM phase modifies low transformer layers far more than seed noise alone (${>}$9${\times}$ in layers 0--7). These changes survive the return to MLM, even when the MLM phase is as long as the CLM phase. Freeze interventions confirm this causally: the downstream benefit requires low-layer modification during CLM, and disappears entirely when these layers are held fixed (\S\ref{sec:analysis}).

The main contributions of this paper are:
\begin{enumerate}
    \item A CLM detour recipe for domain-adaptive encoder pretraining, producing state-of-the-art biomedical encoders in English and French. We release ModernCamemBERT-bio and ModernBERT-bio in Base and Large sizes.

    \item Evidence that the CLM phase leaves lasting changes in low transformer layers that MLM does not reverse, with divergence scaling with model capacity.

    \item Causal evidence via freeze interventions: low layers are necessary for the CLM benefit, mid layers are not.

    \item A practical guideline: 10\% of the CLM budget suffices for the MLM return, confirmed at two scales.
\end{enumerate}

% Figure 1: (left) Pipeline schema, (right) Freeze results bar plot

\section{Related Work}
\label{sec:related_work}

\subsection{Continued Pretraining and Biomedical Encoders}

Domain-adaptive continued pretraining extends general-purpose language models to specific domains by training further on domain-specific corpora. \citet{gururangan2020dont} show that this helps most when the target domain is distant from the pretraining distribution, with biomedical text showing the largest gains. A central debate in biomedical NLP is whether to continue from a general checkpoint or train from scratch on domain data. BioBERT \citep{lee2020biobert} and Bio\_ClinicalBERT \citep{alsentzer2019publicly} take the continued pretraining route from BERT, while PubMedBERT \citep{gu2021pubmedbert} and SciBERT \citep{beltagy2019scibert} train from scratch with domain-specific vocabularies. \citet{gu2021pubmedbert} argue that vocabulary mismatch is the main bottleneck of continued pretraining.

All these models share BERT's 512-token context, which truncates long clinical documents such as discharge summaries or oncology reports. BioClinical-ModernBERT \citep{sounack2025bioclinical} and Clinical ModernBERT \citep{lee2025clinical} address this with the ModernBERT architecture \citep{warner2024modernbert}, supporting 8,192 tokens; BioClinical-ModernBERT trains on 53B tokens in two phases (30\% then 15\% MLM masking). For French, the same debate arises: DrBERT \citep{labrak2023drbert} was pretrained from scratch on 7GB of medical text, while CamemBERT-bio \citep{touchent2024camembertbio} showed that continued pretraining of CamemBERT \citep{martin2020camembert} on a smaller corpus achieves competitive results at a fraction of the cost. Both are limited to 512 tokens. ModernCamemBERT \citep{antoun2025moderncamembert} extends the ModernBERT architecture to French. All of the above use masked language modeling exclusively; none explore alternative training objectives for domain adaptation.

\subsection{CLM and Hybrid Objectives for Encoder Training}

\citet{gisserot2025clm} pretrain encoders from scratch (210M--1B parameters, 100B tokens) and find that a biphasic CLM-then-MLM schedule outperforms pure MLM under fixed compute, with CLM converging faster in early training and producing models that are less sensitive to fine-tuning hyperparameters. However, switching objectives does not always help. \citet{ettin2025} train matched encoder/decoder pairs (up to 1B parameters, 1.7T tokens) and show that continued pretraining on the reverse objective does not bridge the encoder-decoder performance gap, even after 50B tokens of adaptation using masked next-token prediction (MNTP) for the decoder-to-encoder direction. AntLM \citep{antlm2024} takes a different approach, alternating between CLM and MLM epochs while switching both the attention mask and the training objective, and reports gains in both encoder (+2.2pp) and decoder (+1.0pp) directions at small scale (10M words). None of these works analyze why objective switching helps.

\subsection{Representation Similarity and Training Dynamics}

Centered Kernel Alignment \citep[CKA;][]{kornblith2019similarity} compares the internal representations of two networks layer by layer, providing a measure of how similarly they encode the same inputs. CKA has become a standard tool for analyzing how training changes representations in NLP models \citep{wu2020similarity}. \citet{merchant2020what} use CKA to show that task fine-tuning primarily modifies the top layers of BERT while lower layers remain stable.

More broadly, deep networks exhibit critical learning periods where early training conditions leave lasting traces \citep{achille2019critical}. \citet{neyshabur2020being} show that transfer learning benefits concentrate in lower layers, which carry reusable features across tasks. Loss of plasticity can prevent models from adapting to new distributions during continued training \citep{dohare2024loss,ke2022continual}. Layer-freezing interventions \citep{lee2019freezing} provide a tool for establishing which layers causally drive a given effect.

\section{Method}
\label{sec:method}

We compare standard MLM continued pretraining against a two-phase pipeline: CLM detour followed by MLM decay (Figure~\ref{fig:overview}a).

\subsection{Models}

All encoder models use the ModernBERT architecture \citep{warner2024modernbert}, which combines FlashAttention \citep{dao2022flashattention}, rotary positional embeddings \citep{su2024roformer}, alternating local/global attention, and unpadding for 8,192-token sequences. We use two sizes: Base (22 layers, 768 hidden, 12 heads, ${\sim}$150M parameters) and Large (28 layers, 1024 hidden, 16 heads, ${\sim}$350M parameters). For French we start from ModernCamemBERT \citep{antoun2025moderncamembert}; for English from ModernBERT \citep{warner2024modernbert}. As a decoder control (\S\ref{sec:asymmetry}), we use Gemma-3 (270M) \citep{team2025gemma3}. To train this decoder with MLM, we remove the causal attention mask, add a \texttt{<mask>} token to its vocabulary, and train with 30\% masking using the same language model head without the autoregressive position shift. All weights carry over when restoring the causal mask for decay.\label{sec:gemma}

\subsection{Training Pipeline}

The CLM detour consists of two phases. In Phase~1, we replace the bidirectional attention mask with a causal mask and train with next-token prediction. In Phase~2 (decay), we restore bidirectional attention and train with MLM at 15\% masking (the original pretraining rate of ModernBERT) for 10\% of the Phase~1 budget. The optimizer state is kept between phases; only the learning rate scheduler resets. The model architecture is identical between CLM and MLM: only the attention mask (causal vs.\ bidirectional) and loss computation (all tokens vs.\ masked tokens) differ. Phase~2 decays the learning rate from peak to 10\% of peak following the $1{-}\sqrt{t/T}$ schedule of \citet{warner2024modernbert}, without warmup.

The MLM baseline follows the same two-phase structure with 30\% masking in Phase~1 (following \citealp{warner2024modernbert}) and 15\% in Phase~2, identical schedule and optimizer. The only difference is the Phase~1 objective (CLM vs.\ MLM).

\paragraph{Data.} For French, we compile 10B tokens from four sources. The main source (7B tokens) is French biomedical literature (scientific articles, clinical guidelines, and medical theses), where each paragraph is scored for educational value and content richness using an LLM (Qwen3-235B), and articles are upsampled based on their proportion of high-scoring paragraphs, following FineWeb-Edu \citep{penedo2024fineweb} and Biomed-Enriched \citep{touchent2025biomed}. The remaining sources are synthetic medical QA from French coding systems (2B), clinical cases from the European Clinical Case Corpus \citep[E3C;][]{magnini2020e3c} (400M), and drug package inserts from the European Medicines Agency (600M). For English at the 50B scale, we mix biomedical literature from Biomed-Enriched \citep{touchent2025biomed} (60\%, PMC Open Access articles filtered by educational value), medical instruction-following datasets (20\%), and MIMIC-III clinical notes (20\%), trained for a single epoch. A smaller 10B English variant uses Biomed-Enriched with clinical upsampling (80\%) and medical instructions (20\%), without MIMIC.

\paragraph{Training details.} French Base trains for 10B tokens in Phase~1 and 1B in decay; French Large for respectively 25B and 2.5B. English Base is trained at two scales (10B and 50B Phase~1, with proportional decay), and English Large at 50B. All runs use decoupled AdamW with peak lr $2 \times 10^{-4}$, $\beta_1 = 0.9$, $\beta_2 = 0.98$, weight decay $10^{-5}$, and a global batch size of 384 sequences (${\sim}$3.1M tokens). Phase~1 uses linear warmup over 100M tokens then constant learning rate. Documents are packed into 8,192-token sequences with end-of-sequence tokens between documents; attention is not masked across document boundaries. Training uses bf16 mixed precision on 4$\times$H100 GPUs with Composer \citep{mosaicml2022composer}.

\subsection{Freeze Interventions}
\label{sec:freeze}

We run three freeze experiments on the 22-layer French Base model (10B CLM phase, 1B decay), where the CLM-MLM gap is largest (+2.8pp), to test which layers carry the CLM benefit. In each experiment, a contiguous block of layers has its parameters frozen (gradients zeroed, parameters unchanged) during either the CLM phase or the decay phase, while remaining layers train normally. We split the 22 layers into low (0--7) and mid (8--14), approximately the first and second thirds of the network.

\begin{itemize}
    \item \textbf{Experiment~1} (low layers freeze, CLM phase): Layers 0--7 frozen during the CLM phase, then normal decay. Tests whether allowing modifications on low layers during CLM is necessary for the downstream benefit.
    \item \textbf{Experiment~2} (low layers freeze, decay phase): Normal CLM phase, then layers 0--7 frozen during decay. Tests whether low-layer CLM changes persist through decay even without further updates.
    \item \textbf{Experiment~3} (mid layers freeze, CLM phase): Layers 8--14 frozen during the CLM phase, then normal decay. Together with Experiment~1, this tests selectivity: if freezing low layers eliminates the CLM benefit while freezing mid layers preserves it, the effect specifically requires low-layer modifications.
\end{itemize}

The freeze is implemented by zeroing gradients for the specified layers after each backward pass.

\subsection{CKA Methodology}
\label{sec:cka_method}

We measure representational similarity with linear Centered Kernel Alignment \citep[CKA;][]{kornblith2019similarity}. CKA measures how similar two sets of representations are: 1 means identical structure, 0 means no linear relationship. We compute layer-by-layer CKA between model pairs and report divergence ($1 - \text{CKA}$), so that higher values indicate greater representational difference. All CKA computations use float64 arithmetic.
For French, we use 500 held-out texts drawn from the DiaMED clinical case corpus and the FrACCO oncology report corpus (both described in \S\ref{sec:eval}); for English, we use PubMed abstracts. Results are averaged over 3 random seeds (42, 43, 44) for data sampling. To isolate CLM-specific changes from noise introduced by training stochasticity, we compute a seed-noise control: two MLM models trained with different random seeds (17 and 42) but identical data order, so they differ only in dropout and masking patterns. Any divergence exceeding this control can be attributed to the training objective rather than to stochastic variation.

\subsection{Evaluation Protocol}
\label{sec:eval}

We evaluate on 8 French and 11 English biomedical tasks (Table~\ref{tab:eval-tasks} in Appendix~\ref{sec:eval_tasks}), using 9 seeds (42--50) for French and 5 for English. All results use macro-averaged F1 per task, averaged across seeds. French baselines include ModernCamemBERT \citep{antoun2025moderncamembert}, DrBERT \citep{labrak2023drbert}, CamemBERT-bio \citep{touchent2024camembertbio}, and CamemBERT \citep{martin2020camembert}. English baselines include PubMedBERT \citep{gu2021pubmedbert}, BioBERT \citep{lee2020biobert}, SciBERT \citep{beltagy2019scibert}, and BioClinical-ModernBERT \citep{sounack2025bioclinical}.

\section{Experiments}
\label{sec:experiments}

\subsection{French Biomedical Evaluation}

\begin{table*}[t]
\centering
\caption{French biomedical downstream results (macro F1, 9 seeds each). \textbf{Bold}: best per column; \underline{underline}: second best.}
\label{tab:french-results}
\small
\resizebox{\textwidth}{!}{
\begin{tabular}{lcccccccccc}
\toprule
& & \multicolumn{5}{c}{\textbf{Multilabel Classification}} & {\textbf{CLS}} & \multicolumn{2}{c}{\textbf{NER}} & \\
\cmidrule(lr){3-7} \cmidrule(lr){8-8} \cmidrule(lr){9-10}
& \textbf{Ctx} & \textbf{FrACCO-30} & \textbf{FrACCO-100} & \textbf{CANTEMIST} & \textbf{DISTEMIST} & \textbf{MedDialog} & \textbf{DiaMED} & \textbf{EMEA} & \textbf{Medline} & \textbf{Avg} \\
\midrule
\rowcolor{TableSeparator} \multicolumn{11}{c}{\rule{0pt}{10pt}\textbf{Baselines}} \\[1pt]
ModernCamemBERT & 8192 & 70.1 & 55.3 & 63.3 & 20.2 & 60.6 & 56.4 & 68.0 & 59.7 & 56.7 \\
DrBERT & 512 & 53.0 & 35.6 & 37.9 & 21.4 & \underline{63.6} & 57.0 & 69.6 & 62.8 & 50.1 \\
CamemBERT-bio & 512 & 41.9 & 20.1 & 12.8 & 9.6 & 38.6 & 47.7 & \textbf{70.8} & \textbf{65.2} & 38.3 \\
CamemBERT & 512 & 40.8 & 19.4 & 11.9 & 9.5 & 37.4 & 40.6 & 69.5 & 62.7 & 36.5 \\
\rowcolor{TableSeparator} \multicolumn{11}{c}{\rule{0pt}{10pt}\textbf{Our Models (Base, 150M)}} \\[1pt]
MLM baseline & 8192 & 69.9 & 56.8 & 64.9 & 23.5 & 62.5 & 63.4 & 68.5 & 61.4 & 58.9 \\
CLM detour & 8192 & 74.8 & 60.1 & 71.0 & 25.5 & \underline{63.6} & \textbf{67.4} & 68.6 & 61.9 & 61.6 \\
\rowcolor{TableSeparator} \multicolumn{11}{c}{\rule{0pt}{10pt}\textbf{Our Models (Large, 350M)}} \\[1pt]
MLM baseline & 8192 & \underline{79.4} & \underline{63.3} & \underline{72.6} & \underline{29.1} & \textbf{64.5} & 61.2 & \underline{70.4} & \underline{63.5} & \underline{63.0} \\
CLM detour & 8192 & \textbf{80.7} & \textbf{65.4} & \textbf{74.4} & \textbf{30.4} & \textbf{64.5} & \underline{64.8} & 70.3 & 63.1 & \textbf{64.2} \\
\bottomrule
\end{tabular}
}
\end{table*}

Table~\ref{tab:french-results} presents the French results. For Base, the CLM detour achieves 61.6\% average F1, outperforming the MLM baseline on all 8 tasks (+2.8pp). For Large, CLM reaches 64.2\% versus 63.0\% for MLM (+1.2pp). CamemBERT-bio and CamemBERT average 38\% and 37\% overall, limited by their 512-token context on long clinical documents. We release the CLM-detour models as \textbf{ModernCamemBERT-bio} (Base and Large).

\subsection{English Biomedical Evaluation}

\begin{table*}[t]
\centering
\caption{English biomedical results across 11 benchmarks (5 seeds each). Base models trained at 10B and 50B token scales; Large at 50B. \textbf{Bold}: best per column; \underline{underline}: second best. Task abbreviations in Table~\ref{tab:eval-tasks}.}
\label{tab:english-results}
\small
\resizebox{\textwidth}{!}{
\sisetup{detect-weight=true}
\begin{tabular}{ll S[table-format=2.1] S[table-format=2.1] S[table-format=2.1] S[table-format=2.1] S[table-format=2.1] S[table-format=2.1] S[table-format=2.1] S[table-format=2.1] S[table-format=2.1] S[table-format=2.1] S[table-format=2.1] S[table-format=2.1]}
\toprule
& & \multicolumn{5}{c}{\textbf{Clinical Tasks}} & \multicolumn{6}{c}{\textbf{BigBIO Tasks}} & \\
\cmidrule(lr){3-7} \cmidrule(lr){8-13}
& \textbf{Ctx} & \textbf{ChemPr} & \textbf{Pheno} & \textbf{COS} & \textbf{Social} & \textbf{DEID} & \textbf{AnatEM} & \textbf{BC5CDR} & \textbf{JNLPBA} & \textbf{NCBI} & \textbf{GAD} & \textbf{HoC} & \textbf{Avg} \\
& & {\scriptsize Cls} & {\scriptsize Cls} & {\scriptsize NER} & {\scriptsize NER} & {\scriptsize NER} & {\scriptsize NER} & {\scriptsize NER} & {\scriptsize NER} & {\scriptsize NER} & {\scriptsize Cls} & {\scriptsize Cls} & \\
\midrule
\rowcolor{TableSeparator} \multicolumn{14}{c}{\rule{0pt}{10pt}\textbf{Baselines}} \\[1pt]
ModernBERT-base & 8192 & 89.5 & 48.4 & 94.0 & 53.1 & 78.3 & 77.2 & 87.9 & 74.3 & 77.7 & 76.8 & 66.6 & 74.9 \\
PubMedBERT & 512 & 90.2 & 52.0 & 95.0 & 48.7 & 80.4 & \bfseries 83.3 & \underline{89.7} & 74.9 & \bfseries 82.1 & \underline{79.3} & \bfseries 71.0 & 77.0 \\
BioClinical-ModernBERT & 8192 & 90.0 & 60.7 & 94.8 & \underline{56.0} & 81.8 & 79.2 & 88.7 & 74.8 & 78.7 & 75.8 & 67.0 & 77.0 \\
\rowcolor{TableSeparator} \multicolumn{14}{c}{\rule{0pt}{10pt}\textbf{Our Models --- Base 150M (10B tokens)}} \\[1pt]
MLM baseline & 8192 & 90.4 & 61.3 & 94.4 & 55.5 & 79.3 & 78.9 & 88.3 & 74.9 & 79.2 & 78.6 & 68.9 & 77.3 \\
CLM detour & 8192 & 90.3 & 61.0 & 94.3 & 55.7 & 82.6 & 79.9 & 89.1 & 74.2 & 80.4 & \underline{79.3} & 69.4 & 77.8 \\
\rowcolor{TableSeparator} \multicolumn{14}{c}{\rule{0pt}{10pt}\textbf{Our Models --- Base 150M (50B tokens)}} \\[1pt]
MLM baseline & 8192 & \bfseries 90.5 & \underline{61.6} & \underline{95.1} & 55.7 & 81.6 & 80.0 & 88.7 & 74.9 & 80.3 & 78.7 & 67.1 & 77.7 \\
CLM detour & 8192 & 90.1 & \bfseries 61.9 & \bfseries 95.2 & 54.2 & \underline{83.2} & 81.0 & 89.1 & 74.5 & 80.1 & 78.8 & \underline{70.0} & \underline{78.0} \\
\rowcolor{TableSeparator} \multicolumn{14}{c}{\rule{0pt}{10pt}\textbf{Our Models --- Large 350M (50B tokens)}} \\[1pt]
MLM baseline & 8192 & \bfseries 90.5 & 61.0 & 94.9 & 55.0 & 82.3 & 82.0 & 89.4 & \bfseries 75.5 & \underline{81.8} & 76.4 & 67.8 & 77.9 \\
CLM detour & 8192 & 90.4 & 61.3 & 94.7 & \bfseries 56.5 & \bfseries 84.2 & \underline{83.2} & \bfseries 89.8 & \underline{75.3} & 81.7 & \bfseries 79.7 & 69.3 & \bfseries 78.7 \\
\bottomrule
\end{tabular}
}
\end{table*}

Table~\ref{tab:english-results} shows the English results at three scales (Base 10B, Base 50B, and Large 50B). CLM outperforms MLM on average, with the gap widening at Large scale (+0.8pp, 7/11 task wins) compared to Base 10B (+0.5pp) and Base 50B (+0.3pp). The English effect is smaller than in French, with CLM winning 7 of 11 tasks at each Base scale. Baselines include ModernBERT-base (our starting checkpoint), PubMedBERT (512 context), and BioClinical-ModernBERT (8192, standard MLM CPT). PubMedBERT scores higher on short-context BigBIO NER tasks where full-PubMed pretraining helps, but scores 52\% on Phenotype (long-context) versus 61\% for our models.

The smaller English gain is expected. The CLM detour works by reshaping low layers to encode domain-specific features (\S\ref{sec:freeze_results}); when the base model has already seen biomedical text during pretraining, there is less to reshape. ModernBERT was pretrained on web documents, code, and scientific literature following standard modern data mixtures \citep{warner2024modernbert}, which commonly include biomedical corpora such as PubMed. Its low layers already partially encode biomedical features, leaving less room for the CLM detour to help. ModernCamemBERT, by contrast, was pretrained on general French web text without biomedical sources \citep{antoun2025moderncamembert}, so the domain gap is larger and the CLM benefit correspondingly stronger (+2.8pp vs +0.3pp). This suggests that the CLM detour would show larger gains on any domain absent from the base model's pretraining data. We release the English models as \textbf{ModernBERT-bio}.

\subsection{Optimal Decay Ratio}
\label{sec:decay_ratio}

We sweep the decay length from 2.5\% to 50\% of the CLM phase at two scales (Base 10B and Large 25B, evaluated on 3 French tasks). At both scales, 10\% decay gives the best performance; shorter decay (2.5--4\%) scores 1.0pp below optimal and longer decay (20--50\%) provides no additional benefit (Table~\ref{tab:decay_ratio} in Appendix~\ref{app:decay}). This ratio matches the cooldown lengths used in language model pretraining \citep{hu2024minicpm,hagele2024scaling}.

\subsection{Position and Length Analysis}
\label{sec:needle}

Long-context encoding is especially important in biomedical NLP, where clinical documents such as electronic health records, discharge summaries, and oncology reports routinely span thousands of tokens, and where key information (diagnoses, coding labels) can appear anywhere in the document. The CLM detour also improves how the encoder integrates information across long documents. During MLM decay, only 15\% of positions receive a training signal; positions that are rarely masked may be poorly encoded. CLM, which trains on every position, should produce representations that are more uniform across the sequence. We test this with a needle-in-haystack evaluation: we insert a synthetic French medical fact into a biomedical document at a controlled position (start, middle, end) and length (512--8192 tokens), and freeze each encoder to probe whether the fact can be detected from the CLS representation (binary accuracy; details in Appendix~\ref{app:needle}).

\begin{figure}[t]
\centering
\includegraphics[width=\columnwidth]{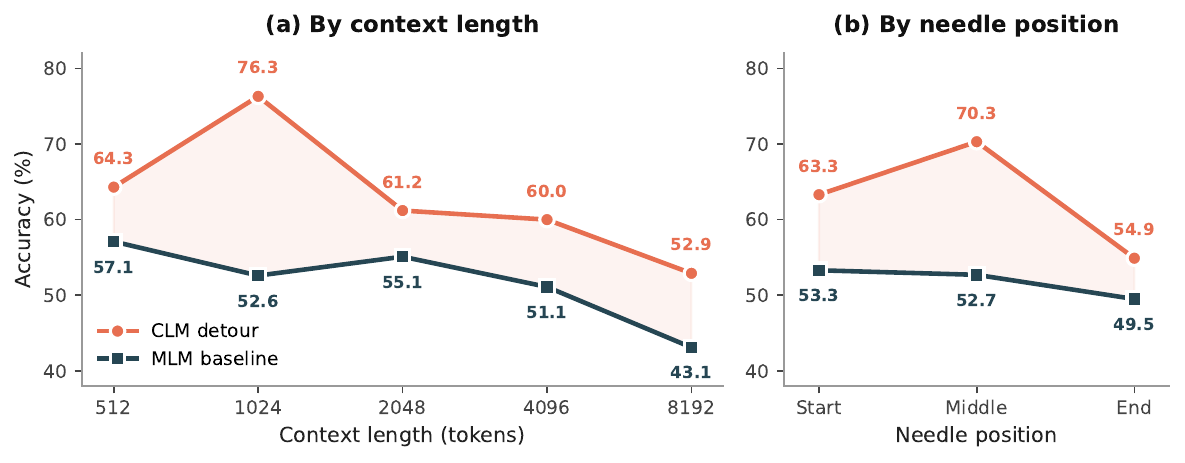}
\caption{Needle-in-haystack evaluation. CLM outperforms MLM at all context lengths and needle positions. Overall: CLM 62.2\% vs MLM 51.6\% (+10.7pp).}
\label{fig:needle}
\end{figure}

CLM outperforms MLM at every context length and position (+10.7pp overall, Figure~\ref{fig:needle}). Two patterns stand out. First, MLM accuracy degrades monotonically with document length (57\% at 512 tokens $\to$ 43\% at 8192), while CLM degrades more slowly and remains above MLM throughout, indicating that CLM representations retain information better over long distances. Second, the CLM advantage is largest at mid-document positions, where the inserted fact is farthest from both the start and end of the sequence. This is the hardest retrieval setting, because the CLS token must integrate information from the middle of a long context, and it is where the difference in position-level training signal matters most. These results are consistent with the downstream pattern: the French tasks with the largest CLM gains (FrACCO, CANTEMIST) require sequence lengths of 4096 tokens during fine-tuning (Table~\ref{tab:eval-tasks}), indicating that their input documents are long enough for mid-document context integration to matter.

\section{Analysis}
\label{sec:analysis}

We now investigate why the CLM detour improves downstream performance, using the CKA methodology described in \S\ref{sec:cka_method}.

\subsection{The CLM Imprint Persists}
\label{sec:persistence}

\begin{figure}[t]
\centering
\includegraphics[width=\columnwidth]{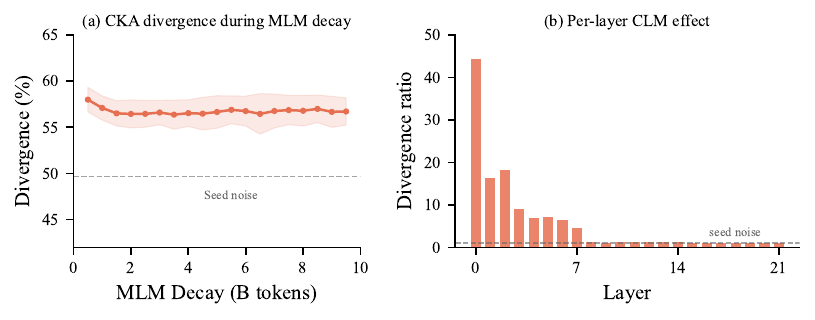}
\caption{(a)~CKA divergence between CLM and MLM models during decay (Base 150M). Dashed line: seed-noise baseline. (b)~Ratio $r_l$ of CLM-MLM divergence to seed-noise divergence for each layer. A ratio of 1 means no CLM-specific effect.}
\label{fig:cka}
\end{figure}

The CLM phase modifies low-layer representations in a way that subsequent MLM training does not undo, even when the decay budget matches the CLM phase. We measure it via CKA divergence between matched CLM-detour and MLM-only models after identical decay, comparing against a seed-noise control: two MLM models trained with different random seeds on identical data and hyperparameters.

The CLM imprint is lasting. CKA divergence reaches ${\sim}$56.5\% after 1.5B tokens of MLM decay and remains stable, with only 0.6pp variation over 8B additional tokens (Figure~\ref{fig:cka}a). The seed-noise control shows 49.7\% divergence overall, confirming that CLM adds signal beyond stochastic training variation.

The CLM imprint concentrates in low layers. Both CLM and MLM continued pretraining modify all layers relative to the starting checkpoint, and mid/deep layers diverge heavily under either objective (Appendix~\ref{app:cka-raw}). To isolate where CLM differs from MLM \emph{specifically}, we normalize each layer's CLM-MLM divergence by the seed-noise baseline (Figure~\ref{fig:cka}b):
\[
r_\ell = \frac{1 - \text{CKA}(\text{CLM}_\ell,\; \text{MLM}^{s_1}_\ell)}{1 - \text{CKA}(\text{MLM}^{s_2}_\ell,\; \text{MLM}^{s_1}_\ell)}
\qquad (s_1, s_2\text{: different random seeds})
\]
A ratio of 1 means no CLM-specific effect at that layer. Low layers (0--7) show ratios of 5--44${\times}$: CLM changes them far more than random seed variation does. Mid and deep layers are near 1${\times}$: both objectives modify them similarly. We test whether these low-layer changes are the causal driver of downstream improvement in \S\ref{sec:freeze_results}.

\subsection{Causal Evidence: Freeze Interventions}
\label{sec:freeze_results}

Do these low-layer changes actually drive the downstream improvement? We answer with the freeze interventions described in \S\ref{sec:freeze} (Figure~\ref{fig:overview}b; full results in Table~\ref{tab:freeze} in Appendix~\ref{app:freeze}).

Freezing low layers (0--7) during CLM drops F1 from 61.6\% to 59.3\%. We cannot reject the null hypothesis that the resulting model matches the MLM baseline ($p = 0.25$, paired bootstrap across 8 tasks $\times$ 9 seeds). Freezing mid layers (8--14) during CLM preserves the benefit (61.2\%). Together, these two experiments show that low layers are necessary and mid layers are not. Freezing low layers during decay has negligible effect ($-$0.4pp, 61.2\%), confirming that MLM decay already preserves the CLM imprint in low layers. These experiments use French Base models. The CKA patterns that motivate them (low-layer divergence well above seed noise) are consistent across French Base, French Large, and English Base (Table~\ref{tab:asymmetry}).

\subsection{Non-Localizability}

The freeze experiments show that low layers are necessary \emph{during training}. We now ask whether the resulting benefit can be localized \emph{after training}. We copy full parameter blocks (self-attention, MLP, and layer norms) for a group of layers from the CLM model into the MLM model, keeping embeddings and remaining layers from MLM, and evaluate via linear probing on frozen representations (logistic regression, 3 seeds). We use DiaMED, the task with the largest CLM-MLM gap (7.1pp), for clearer signal.

\begin{table}[t]
\centering
\caption{Layer transplantation on DiaMED (3 seeds, linear probing). No transplant recovers more than 35\% of the 7.1pp gap.}
\label{tab:transplants}
\begin{tabular}{lccc}
\toprule
Transplant & F1 (\%) & $\Delta$ & Recovery \\
\midrule
\rowcolor{TableSeparator} \multicolumn{4}{c}{\rule{0pt}{9pt}\textbf{References}} \\[1pt]
MLM baseline & 32.8 & --- & 0\% \\
CLM detour & 39.9 & +7.1 & 100\% \\
\midrule
\rowcolor{TableSeparator} \multicolumn{4}{c}{\rule{0pt}{9pt}\textbf{Layer group transplants}} \\[1pt]
Low (0--7) & 31.9 & $-$0.9 & $-$12\% \\
Mid (8--14) & 35.3 & +2.5 & +35\% \\
Late (15--21) & 32.4 & $-$0.4 & $-$6\% \\
\bottomrule
\end{tabular}
\end{table}

No transplant recovers more than 35\% of the gap (Table~\ref{tab:transplants}). Low and late layer transplants hurt performance. Inserting CLM low layers into an MLM model creates a mismatch with the mid and deep layers that co-adapted under MLM, yielding representations worse than either pure model. Component-level transplants (attention-only, MLP-only) show a similar pattern (Appendix~\ref{app:transplant-comp}). Unlike freeze interventions, which act during training, transplants replace already-trained weights. The freeze experiments show low layers are necessary during training; the transplant experiments show the resulting benefit cannot be isolated post hoc.

\subsection{Scaling and Architectural Asymmetry}
\label{sec:asymmetry}

\begin{table}[t]
\centering
\caption{CKA divergence by model size and architecture. The encoder-decoder comparison uses different model families (see \S\ref{sec:asymmetry}).}
\label{tab:asymmetry}
\begin{tabular}{llcc}
\toprule
Model & Params & Div. & CI \\
\midrule
\rowcolor{TableSeparator} \multicolumn{4}{c}{\rule{0pt}{9pt}\textbf{Encoders (MLM$\to$CLM$\to$MLM)}} \\[1pt]
ModernCamemBERT Base & 150M & 56.5\% & $\pm$0.7\% \\
ModernCamemBERT Large & 350M & 67.2\% & $\pm$0.3\% \\
\midrule
\rowcolor{TableSeparator} \multicolumn{4}{c}{\rule{0pt}{9pt}\textbf{Decoder (CLM$\to$MLM$\to$CLM)}} \\[1pt]
Gemma-3 & 270M & 26.3\% & $\pm$1.0\% \\
\bottomrule
\end{tabular}
\end{table}

The CLM imprint increases with model capacity (Table~\ref{tab:asymmetry}): Large (350M) reaches 67.2\% $\pm$ 0.3\% divergence versus 56.5\% $\pm$ 0.7\% for Base (150M). We also test the reverse on a decoder: starting from Gemma-3 (270M; \citealp{team2025gemma3}), we remove the causal attention mask and train with MLM for 10B tokens, then restore the causal mask and return to CLM for 1B tokens (details in \S\ref{sec:gemma}). The decoder retains only 26.3\% $\pm$ 1.0\% divergence versus 67.2\% for an encoder of similar scale, a suggestive 2.6${\times}$ asymmetry. One explanation is the density of the decay signal: MLM decay at 15\% masking provides weak erasure, while CLM decay at 100\% provides strong erasure. All confidence intervals are 95\% over 3 seeds. This comparison is exploratory: it is confounded by model family and language, and a controlled comparison using encoder and decoder variants of the same family would be needed to confirm the asymmetry.

\section{Conclusion}
\label{sec:conclusion}

A temporary detour through causal language modeling during encoder continued pretraining produces lasting representational changes that improve downstream performance. The effect is not an artifact of additional compute: matched MLM baselines with identical data and training budget score lower. The CLM imprint concentrates in low transformer layers, persists through extended MLM decay, and is not localized to a single layer group and is not erased by continued training. Freeze interventions show that low-layer changes are necessary for the downstream improvement, while mid-layer changes are not. As a practical guideline, MLM decay of roughly 10\% of the detour length suffices to stabilize the model without erasing the CLM imprint.

These findings suggest that the sequence of training objectives shapes the final model, not just the total compute or the final objective. We speculate that this mechanism extends beyond biomedical text to other domain adaptation settings where the target domain differs substantially from the pretraining distribution. The stronger effect in French (+2.8pp) compared to English (+0.3--0.8pp depending on model size) suggests that lower-resource biomedical languages such as Italian, Spanish, or Portuguese, where less domain text is available during general pretraining, would likely benefit more from the CLM detour.

We release ModernCamemBERT-bio and ModernBERT-bio as open biomedical encoders in Base and Large sizes with 8,192-token context.

% ============================================================================
% BIBLIOGRAPHY
% ============================================================================
\bibliographystyle{plainnat}
\bibliography{references}

% ============================================================================
% LIMITATIONS + APPENDIX
% ============================================================================
\section*{Limitations}

\paragraph{Domain and language scope.} All experiments use biomedical text in French and English. The CLM detour gains are larger in French (+2.8pp, 8/8 task wins) than in English (+0.3--0.8pp, 7/11 task wins), consistent with the larger domain gap for French. The gap narrows from 10B to 50B tokens in English, suggesting that additional data partially closes it. We have not tested whether the effect generalizes to other domains or languages.

\paragraph{Scale and architecture.} Our largest model has 350M parameters. The CLM imprint increases with scale (56.5\% for Base, 67.2\% for Large), but we have not verified this trend at billion-parameter scale. The encoder-decoder asymmetry (2.6${\times}$) compares models that differ in architecture, language, and pretraining data; a controlled comparison using encoder and decoder variants of the same model family would provide stronger evidence.

\paragraph{Freeze experiments.} The freeze interventions were conducted on French Base models only, where the CLM-MLM gap is largest. Replicating on English or Large models would strengthen the generalization of the causal claim. The freeze granularity (7--8 layer blocks) is coarse; finer-grained interventions could narrow down the critical layers further.

\appendix

\section{English Training Data}
\label{app:en-data}

The English biomedical component (60\% of the 50B mix) uses the Biomed-Enriched dataset \citep{touchent2025biomed}, a curated version of PMC Open Access articles where each paragraph is scored for educational value using an LLM classifier. We retain paragraphs scoring $\geq$3.0. For the 10B variant, clinical case paragraphs are upsampled 100$\times$ and other clinical documents 10$\times$ to increase clinical representation.

The instruction-following component (20\%) consists of five datasets: MedS-Ins \citep{wu2025medsins}, ReasonMed \citep{sun2025reasonmed}, MediFlow \citep{corbeil2025mediflow}, UltraMedical \citep{zhang2024ultramedical}, and EHR-Ins-Reasoning \citep{liao2025ehrr1}. The 10B variant uses only MediFlow.

The clinical component (20\%, 50B only) uses MIMIC-III discharge summaries and clinical notes, upsampled across multiple passes to reach the target token count.

\section{Freeze Intervention Results}
\label{app:freeze}

\begin{table}[H]
\centering
\caption{Freeze intervention results (French Base, 8 tasks, 9 seeds).}
\label{tab:freeze}
\begin{tabular}{lcc}
\toprule
Condition & Avg F1 & $\Delta$ vs CLM detour \\
\midrule
Freeze L0--7 (CLM phase) & 59.3 & $-$2.3 \\
Freeze L8--14 (CLM phase) & 61.2 & $-$0.4 \\
Freeze L0--7 (decay phase) & 61.2 & $-$0.4 \\
\bottomrule
\end{tabular}
\end{table}

\section{Component Transplant Results}
\label{app:transplant-comp}

\begin{table}[H]
\centering
\caption{Component-level transplants on DiaMED (3 seeds, linear probing).}
\begin{tabular}{lccc}
\toprule
Transplant & F1 (\%) & $\Delta$ & Recovery \\
\midrule
All attention & 33.9 & +1.1 & +15\% \\
All MLP & 35.3 & +2.5 & +35\% \\
\bottomrule
\end{tabular}
\end{table}

\section{Decay Ratio Sweep}
\label{app:decay}

\begin{table}[H]
\centering
\caption{Decay ratio sweep on 3 French tasks (DiaMED, FrACCO-30, FrACCO-100; 9 seeds each). 10\% decay is optimal at both scales.}
\label{tab:decay_ratio}
\begin{tabular}{lccc}
\toprule
Decay tokens & Decay / Phase~1 & CLM+decay F1 & MLM+decay F1 \\
\midrule
\rowcolor{TableSeparator} \multicolumn{4}{c}{\rule{0pt}{9pt}\textbf{10B Phase~1 (Base, 150M)}} \\[1pt]
0.25B & 2.5\% & 71.6\% & 69.1\% \\
\textbf{1.0B} & \textbf{10\%} & \textbf{72.6\%} & 70.4\% \\
2.0B & 20\% & 71.9\% & 70.4\% \\
3.0B & 30\% & 72.1\% & 69.7\% \\
5.0B & 50\% & 71.7\% & 70.4\% \\
\rowcolor{TableSeparator} \multicolumn{4}{c}{\rule{0pt}{9pt}\textbf{25B Phase~1 (Large, 350M)}} \\[1pt]
1.0B & 4\% & 73.2\% & 70.1\% \\
\textbf{2.5B} & \textbf{10\%} & \textbf{73.5\%} & 70.5\% \\
5.0B & 20\% & 72.7\% & 70.1\% \\
\bottomrule
\end{tabular}
\end{table}

\section{Evaluation Tasks}
\label{sec:eval_tasks}

\begin{table}[H]
\centering
\caption{Evaluation tasks. Max seq.\ length reflects the fine-tuning configuration.}
\label{tab:eval-tasks}
\small
\begin{tabular}{lllcl}
\toprule
\textbf{Task} & \textbf{Type} & \textbf{Lang} & \textbf{Max len.} & \textbf{Reference} \\
\midrule
DiaMED & Classification & FR & 2048 & \citep{labrak2024drbenchmark} \\
MedDialog-FR & Multilabel & FR & 4096 & \citep{liu2024meddialog} \\
FrACCO (30/100) & Multilabel & FR & 4096 & \citep{pignat2025fracco} \\
CANTEMIST & Multilabel & FR & 4096 & \citep{miranda2020cantemist} \\
DisTEMIST & Multilabel & FR & 4096 & \citep{miranda2022distemist} \\
QUAERO (EMEA/Medline) & NER & FR & 1024 & \citep{neveol2014quaero} \\
\midrule
ChemProt & Relation Extr. & EN & 512 & \citep{krallinger2017overview} \\
Phenotyping & Classification & EN & 8192 & \citep{gehrmann2018comparing} \\
COS & NER & EN & 512 & \citep{klassen2014annotating} \\
Social History & NER & EN & 512 & \citep{lybarger2023n2c2} \\
De-identification & NER & EN & 8192 & \citep{neamatullah2008automated} \\
AnatEM & NER & EN & 512 & \citep{pyysalo2014anatomical} \\
BC5CDR & NER & EN & 512 & \citep{li2016biocreative} \\
JNLPBA & NER & EN & 512 & \citep{collier2004introduction} \\
NCBI Disease & NER & EN & 512 & \citep{dogan2014ncbi} \\
GAD & Classification & EN & 512 & \citep{bravo2015extraction} \\
HoC & Classification & EN & 512 & \citep{baker2016automatic} \\
\bottomrule
\end{tabular}
\end{table}

\section{Raw CKA Divergence Curves}
\label{app:cka-raw}

\begin{figure}[H]
\centering
\includegraphics[width=\columnwidth]{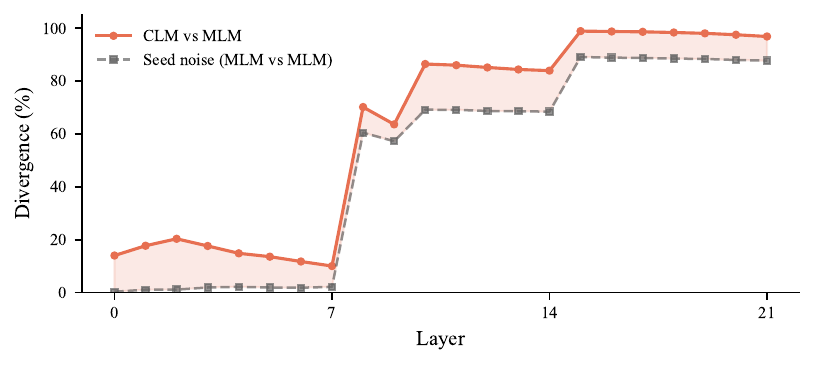}
\caption{Per-layer CKA divergence for CLM vs MLM (coral) and seed noise (gray). Both continued pretraining objectives modify mid and deep layers heavily, but only CLM produces large changes in low layers (0--7). The shaded area is the CLM-specific signal, which the divergence ratio in Figure~\ref{fig:cka}b normalizes.}
\end{figure}

\section{CKA Formula}
\label{app:cka}

We use linear CKA \citep{kornblith2019similarity}. Given two representation matrices $X \in \mathbb{R}^{n \times p}$ and $Y \in \mathbb{R}^{n \times q}$ (one per model, $n$ samples, mean-pooled over tokens), we center each column, compute Gram matrices $K = XX^\top$ and $L = YY^\top$, and define:
\[
\text{CKA}(X, Y) = \frac{\text{tr}(KL)}{\sqrt{\text{tr}(K^2)\,\text{tr}(L^2)}}
\]
We report divergence $d = 1 - \text{CKA}$, so higher values indicate greater representational difference. All computations use float64 arithmetic.

\section{Fine-tuning Protocol}
\label{app:finetuning}

\paragraph{French tasks.} Classification (DiaMED) and multilabel tasks (MedDialog, FrACCO, CANTEMIST, DisTEMIST) use a linear head on the CLS token, trained for 15 epochs with lr $= 2 \times 10^{-5}$, batch size 4, and 10\% warmup. Multilabel tasks use max sequence length 4096 with weighted BCEWithLogitsLoss; classification uses 2048. NER tasks (EMEA, Medline) use a span-based architecture with BiLSTM, CRF, and biaffine scorer, trained for 4000 steps with lr $= 5 \times 10^{-5}$ for the encoder and $10^{-3}$ for the head, batch size 16, max length 1024. All tasks use AdamW with weight decay 0.01 and select the best checkpoint by validation F1.

\paragraph{English tasks.} We follow the BioClinical-ModernBERT evaluation protocol \citep{sounack2025bioclinical}: lr $= 5 \times 10^{-5}$, weight decay 0.01, batch size 16, 10 epochs for most tasks (20 for NER). All models are fine-tuned with the same hyperparameters per task.

\section{Needle-in-Haystack Protocol}
\label{app:needle}

We design a synthetic needle-in-haystack task in French to evaluate long-context retrieval. A medical fact (the ``needle'') is inserted into a clinical document (the ``haystack'') at a controlled position, and the model must determine whether a given query fact is present (binary classification). Negative examples use the same template with different slot values, so the task requires precise retrieval rather than shallow pattern matching.

\paragraph{Fact templates.} We define 8 French medical fact templates covering drugs, vital signs, lab results, diagnoses, procedures, allergies, treatment duration, and symptoms. Each template has 2--4 slots filled from small lexicons. Examples:

\begin{itemize}
  \item \textit{Le patient a re\c{c}u 200\,mg de morphine par voie intraveineuse.}\\
        \textbf{Distractor:} \textit{Le patient a re\c{c}u 500\,mg de tramadol par voie orale.}
  \item \textit{La tension art\'erielle mesur\'ee \'etait de 160/90\,mmHg.}
  \item \textit{Le diagnostic retenu est cirrhose h\'epatique de stade~III.}
  \item \textit{Le patient rapporte une dyspn\'ee \'evoluant depuis une semaine.}
\end{itemize}

\paragraph{Dataset and evaluation.} Haystacks are drawn from French biomedical text. We generate 1500 balanced positive/negative pairs across 5 lengths (512--8192 tokens) and 3 positions (start, middle, end), split 70/15/15 for train/validation/test. We freeze each encoder and train a 2-layer MLP probe (Dropout $\to$ Linear $\to$ GELU $\to$ Dropout $\to$ Linear) on CLS representations for 3 epochs (lr $= 2 \times 10^{-5}$, batch size 4, AdamW), selecting the best checkpoint by validation accuracy.

\end{document}